\documentclass[conference]{IEEEtran}
\IEEEoverridecommandlockouts
\usepackage{cite}
\usepackage{amsmath,amssymb,amsfonts}
\usepackage{algorithmic}
\usepackage{algorithm}
\usepackage{graphicx}
\usepackage{textcomp}
\usepackage{amsthm}
\usepackage{xcolor}
\usepackage{booktabs}
\usepackage{subfig}
\def\BibTeX{{\rm B\kern-.05em{\sc i\kern-.025em b}\kern-.08em
    T\kern-.1667em\lower.7ex\hbox{E}\kern-.125emX}}

\begin{document}

\newcommand{\customoverline}[1]{\mkern 0.5mu \overline{\mkern-3.5mu#1\mkern0.0mu}\mkern 0.5mu}

\title{Informative Subgraphs Aware Masked Auto-Encoder in Dynamic Graphs}

\author{
\IEEEauthorblockN{Pengfei Jiao$^1$, Xinxun Zhang$^1$, Mengzhou Gao$^1$\IEEEauthorrefmark{2}\thanks{† Corresponding author.}, Tianpeng Li$^2$, Zhidong Zhao$^1$}

\IEEEauthorblockA{$^1$ {Hangzhou Dianzi University, Hangzhou, China}}
\IEEEauthorblockA{$^2$ {Tianjin University, Tianjin, China}}
\IEEEauthorblockA{pjiao@hdu.edu.cn, xxzhang@hdu.edu.cn, mzgao@hdu.edu.cn, ltpnimeia@tju.edu.cn, zhaozd@hdu.edu.cn}
}

\maketitle

\begin{abstract}
  Generative self-supervised learning (SSL), especially masked autoencoders (MAE), has greatly succeeded and garnered substantial research interest in graph machine learning. However, the research of MAE in dynamic graphs is still scant. This gap is primarily due to the dynamic graph not only possessing topological structure information but also encapsulating temporal evolution dependency. Applying a random masking strategy which most MAE methods adopt to dynamic graphs will remove the crucial subgraph that guides the evolution of dynamic graphs, resulting in the loss of crucial spatio-temporal information in node representations. To bridge this gap, in this paper, we propose a novel \textbf{I}nformative \textbf{S}ubgraphs Aware Masked Auto-Encoder in \textbf{Dy}namic \textbf{G}raph, namely DyGIS. Specifically, we introduce a constrained probabilistic generative model to generate informative subgraphs that guide the evolution of dynamic graphs, successfully alleviating the issue of missing dynamic evolution subgraphs. The informative subgraph identified by DyGIS will serve as the input of dynamic graph masked autoencoder (DGMAE), effectively ensuring the integrity of the evolutionary spatio-temporal information within dynamic graphs. Extensive experiments on eleven datasets demonstrate that DyGIS achieves state-of-the-art performance across multiple tasks. 
\end{abstract}


\section{Introduction}
Self-supervised learning (SSL) on graph data has emerged as a highly promising learning paradigm due to its powerful modeling capabilities, especially when explicit labels are not available. Currently, most SSL methods on graphs can be mainly divided into two categories \cite{liu2021self}, \textit{i.e.}, contrastive SSL methods, and generative SSL methods. Contrastive SSL methods learn effective node representations by constructing positive and negative sample pairs and minimizing the distance between positive sample pairs while maximizing the distance between negative sample pairs \cite{hassani2020contrastive, zhu2021graph}. Compared with contrastive SSL, generative SSL methods directly reconstruct the input graph data as the training objective without requiring complex graph data augmentations and training strategies, as well as negative sampling \cite{wu2021self}. 

In recent years, graph mask autoencoders (GMAE), a form of generative SSL, have demonstrated excellent performance by reconstructing the masked content from the unmasked part \cite{tian2023heterogeneous,hou2023graphmae2,tan2023s2gae,li2023seegera}. Due to its competitive performance compared to contrastive SSL methods and simple training strategy, GMAE quickly became one of the dominant paradigms in current graph machine learning. For instance, GraphMAE \cite{hou2022graphmae} proposes to reconstruct node features with masking on graphs. MaskGAE \cite{li2023s} reconstructs graph structure with masks to obtain node representations. Although these methods have achieved significant success, they all focus on static graphs and overlook the dynamic nature of real-world graphs. Therefore, exploring how to apply masked autoencoder (MAE) to the dynamic graph is more challenging but of high importance as it describes how the real dynamic system interacts and evolves.


Dynamic graphs are prevalent in the real world. Many real-world situations, such as social networks \cite{yang2020relation}, transportation networks \cite{zhao2019t}, trade networks \cite{liu2019characterizing}, and more, can be modeled as dynamic graphs, where the relationships between nodes and edges evolve over time. Research and applications focused on dynamic graphs are crucial for a better understanding and handling of complex dynamic systems in the real world \cite{zhang2024tifs}. Typically, a dynamic graph can be modeled using a sequence of snapshots in temporal order, where each snapshot portrays the dynamic evolution relationships at the current moment \cite{jiao2022tnnls,jiao2023tkde}. However, modeling the dynamic graph using the MAE framework is a non-trivial task due to the evolutionary nature of dynamic graphs. 

Most current GMAE methods adopt a random masking strategy to mask graph structures or node features \cite{hou2022graphmae,hou2023graphmae2, li2023s,tan2023s2gae}. Intuitively, we can also directly employ a random masking strategy for each snapshot of the dynamic graph. However, the random masking strategy has the following problems:
1) The evolution direction of dynamic graphs is commonly determined by crucial substructures. For example, in a social network, nodes with more friends (\textit{i.e.}, nodes with larger degrees) often play a more decisive role in shaping the development trend of the social network. Therefore, if the random masking strategy removes substructures containing such nodes, it inevitably leads to the loss of spatio-temporal information in dynamic graphs. This issue is particularly evident with a high mask ratio, as random masking with a high mask ratio has a higher probability of dropping more critical edges and nodes.
2) Unlike static graphs, dynamic graphs with multiple snapshots are temporally dependent on each other, so most dynamic graph neural networks (DyGNNs) will pass their hidden state as input for the next step to extract spatio-temporal information \cite{hajiramezanali2019variational, yang2021discrete, bai2023hgwavenet, ltp_tcyb}. Once the critical evolving subgraph of any snapshot is removed, it will lead to the loss of critical information in the hidden state of the current snapshot, affecting the spatio-temporal information extraction of subsequent snapshots. Furthermore, due to the cascade effect of dynamic graphs, it will make DyGNN evolve in the wrong direction. In summary, the random masking strategy leads to the loss of evolving substructures in the dynamic graph, resulting in the latent node representations lacking crucial spatio-temporal information.

To address the issues mentioned above, we define the critical subgraph that guides the evolution of dynamic graphs as the informative subgraph, and the edges within this subgraph as informative edges. Considering the limitations of the random masking strategy in the dynamic graph discussed previously, we argue that a constrained generative model should be imposed to generate informative subgraphs to capture the evolving patterns of the dynamic graph when applying MAE to dynamic graphs. 
From our analysis above, we know that the effective spatio-temporal information in node representations is mainly extracted from the informative subgraphs of the dynamic graph. Removing these subgraphs will result in the loss of spatio-temporal information. Intuitively, when the spatio-temporal information of node representations is disrupted, the informative subgraphs obtained from node representations will not be reconstructed well within the generative SSL framework.
Therefore, we came up with an idea to perturb the crucial spatio-temporal information in the latent representation space of dynamic graphs and then generate informative subgraphs based on the reconstruction performance.

In this paper, we propose a novel \textbf{I}nformative \textbf{S}ubgraphs Aware Masked Auto-Encoder in \textbf{Dy}namic \textbf{G}raph, namely DyGIS. DyGIS introduces additional constraints to ensure the generated informative subgraphs guide the evolution of the dynamic graph, eliminating the effect of missing crucial information due to applying a random masking strategy to the dynamic graph. Specifically, we first introduce noisy random graphs that share the same statistical properties as the dynamic graph. Considering that mutual information can measure the correlation between two variables. Therefore, we maximize the mutual information between the latent representations of these two graphs to perturb informative spatio-temporal information of the node embedding in dynamic graphs. Subsequently, we use the perturbed latent representations to reconstruct dynamic graphs, generating informative subgraphs based on the reconstruction performance. The process of generating subgraphs can be viewed as a masking strategy, as the informative subgraphs serve as the input of dynamic graph masked autoencoder (DGMAE). Our main contributions can be summarized as follows:
\begin{itemize}
    \item We introduce a novel generative model, DyGIS, that successfully applies MAE to dynamic graphs. To the best of our knowledge, we are currently the first to explore generative SSL with masking on dynamic graphs.
    \item We propose an informative subgraph generator to generate informative subgraphs that guide the evolution of dynamic graphs, alleviating the loss of crucial spatio-temporal information and ensuring the integrity of evolution patterns in dynamic graphs.
    \item We perform extensive experiments to demonstrate the effectiveness of our proposed approach, which achieves state-of-the-art performances compared to DyGNN methods and the most relevant GMAE methods across various tasks.
    
\end{itemize}

\section{Related Work}
\subsection{Dynamic Graph Neural Networks}
Incorporating temporal information and graph topology features into graph representation learning has attracted increasing attention since most real-world graphs are dynamic \cite{huang2023TGB, huang2023benchtemp}. Based on the form of dynamic graphs, DyGNN methods can be divided into two main categories: continuous-time DyGNNs and discrete-time DyGNNs \cite{kazemi2020representation}. Continuous-time DyGNNs view a dynamic graph as a flow of edges with specific timestamps \cite{nguyen2018continuous,jin2022neural}. Discrete-time DyGNNs view a dynamic graph as a series of snapshots with timestamps \cite{sankar2020dysat, hajiramezanali2019variational, zhang2022dynamic}. Our work relates to the second category, discrete-time DyGNNs. 

The essence of both DyGNNs is to encode the temporal dependencies and spatial information of dynamic graphs into node embedding to support various downstream tasks such as link prediction and node classification. Specifically, VGRNN \cite{hajiramezanali2019variational} introduces additional latent random variables to capture both topology and time attributes. DySAT \cite{sankar2020dysat} obtains node embeddings through a self-attentive mechanism that unites structural neighborhoods and temporal dynamics. EvolveGCN \cite{pareja2020evolvegcn} captures the dynamics of graph sequences by using an RNN to update the GCN parameters. HTGN \cite{yang2021discrete}and HWaveNet \cite{bai2023hgwavenet} utilize hyperbolic geometry to learn the spatial topological structures and temporal evolutionary information. Meanwhile, graph contrastive learning, a powerful self-supervised method, has been successful in dynamic graphs.  DGCN \cite{gao2023novel} maximizes the mutual information between the local and global representations of graphs at each time step. 
DyTed \cite{zhang2023dyted} introduces temporal-invariant and structure-proximity contrastive learning to obtain disentangled representation. 
\subsection{Graph Masked Autoencoders}
GMAE is a generative SSL method that has very recently been extensively studied due to its simple design and excellent performance. The essence of GMAE is to force the model to learn more critical underlying patterns by masking part of the graph data. GraphMAE \cite{hou2022graphmae} and RARE \cite{tu2023rare} concentrate on reconstructing the feature matrices with masking to obtain node embeddings. On the other hand, MaskGAE \cite{li2023s} and S2GAE \cite{tan2023s2gae} focus on reconstructing masked graphs and systematically analyzing the underlying reasons for GMAE's effectiveness. GraphMAE2 \cite{hou2023graphmae2} introduces more powerful GNN-based decoders to enhance the model's learning capability. SeeGera \cite{li2023seegera} and GiGaMAE \cite{shi2023gigamae} emphasize feature reconstruction and structural masking with the variational framework and mutual information respectively. HGMAE \cite{tian2023heterogeneous} extends MAE to heterogeneous graphs to capture comprehensive graph information. In addition, GMAE has been applied in various fields such as recommendation systems \cite{10.1145/3539618.3591692} and molecular modeling \cite{liu2023rethinking}. Despite the promising results of all the above work in GMAE, attempts in dynamic graphs are still scant.

\begin{figure*}[t]
\centering
\includegraphics[width=1.0\textwidth]{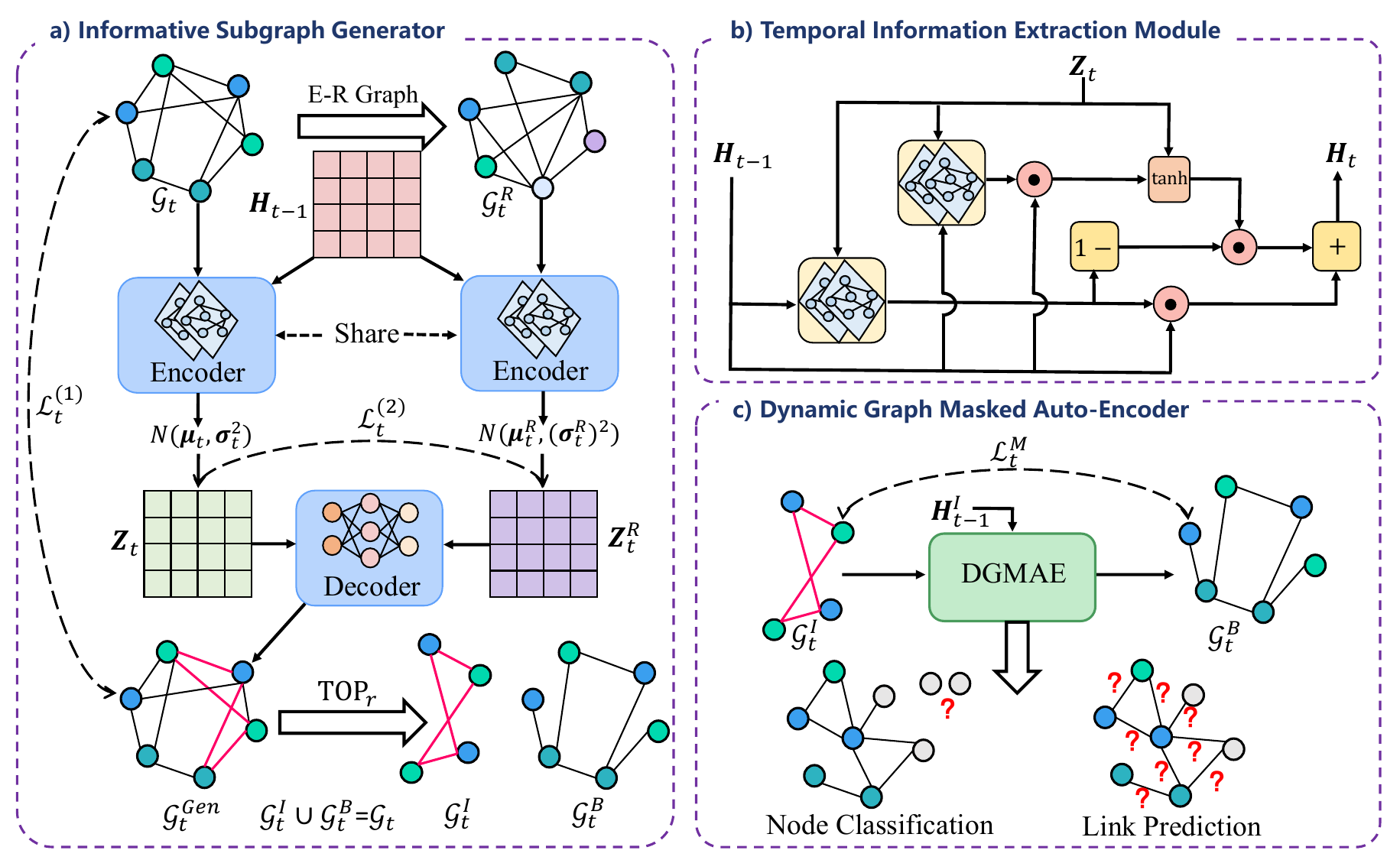} 
\caption{ The overall framework of the DyGIS consists of three components. a) Informative Subgraph Generator: it's used to generate informative subgraph $\mathcal{G}_t^{I}$ and bias subgraph $\mathcal{G}_t^{B}$ via adaptive learning. b) Temporal Information Extraction Module: utilized to update and pass temporal dependencies between dynamic graphs. c) Dynamic Graph Masked Auto-Encoder: $\mathcal{G}_t^{I}$ serves as its input and then reconstructs $\mathcal{G}_t^{B}$ to obtain node representation applied to downstream tasks.}
\label{fig2}
\end{figure*}

\section{Preliminaries}
\subsection{Mutual Information}
In this paper, we resort to mutual information to perturb the informative spatio-temporal information of node embedding and generate informative edges since it can measure the dependency relationship between variables \cite{gabrie2018entropy}. The mutual information between two variables $X, Y$ is formalized as,
\begin{equation}
    I(X;Y) = \sum_{x \in X} \sum_{y \in Y} {\rm log}(\frac{p(x,y)}{p(x)p(y)})
\end{equation}
where $p(x, y)$ is the joint probability distribution function of $X$ and $Y$, and $p(x)$ and $p(y)$  are the marginal probability distribution functions of $X$ and $Y$, respectively. 

However, directly calculating mutual information is not straightforward due to the unknown distribution of variables. Therefore, some methods have emerged to estimate mutual information \cite{poole2019variational}. We adopt InfoNCE \cite{oord2018representation} to estimate mutual information. On one hand, InfoNCE has been widely applied and proven effective in various contexts \cite{shi2023gigamae, tschannen2019mutual,zhang2023dyted}. On the other hand, some research has demonstrated that InfoNCE serves as a lower bound for mutual information \cite{oord2018representation}. We can maximize the InfoNCE loss to maximize the mutual information between variables. Specifically, given a pair of node embeddings $(p_i, q_i)$, the InfoNCE loss can be formalized as, 
\begin{equation}
    \mathcal{L}^{\mathcal{D}}(p_i, q_i) = {\rm log} \frac{\mathcal{D}(p_i, q_i)}{\sum_{j=1}^{N} \mathcal{D}(p_i, q_j) + \sum_{j=1}^{N}\mathcal{D}(p_i, p_j) - \mathcal{D}(p_i, p_i) }
    \label{eq2}
\end{equation}
where $\mathcal{D}$ is the discriminator function, and $N$ is denote the number of negative samples. In this paper, we define anchor pairs between the dynamic graph and the noisy random graph as positive pairs, while the rest are considered negative pairs. 

\subsection{Problem Definition}
In this paper, we focus on discrete-time DyGNNs. A dynamic graph can be viewed as a series of snapshots $G=\left\{\mathcal{G}_1, \mathcal{G}_2,\cdots, \mathcal{G}_T\right\}$, where $T$ is the total number of snapshots. $\mathcal{G}_t$ denotes the snapshot at time $t$, which is a graph with a node-set $\mathcal{V}_t$, an edge-set $\mathcal{E}_t$ and feature matrix $ \mathcal{X}_t$, \textit{i.e.}, $\mathcal{G}_t=(\mathcal{V}_t, \mathcal{E}_t,  \mathcal{X}_t)$. We denote $\mathcal{A}_t$ and as the adjacency matrix of $\mathcal{G}_t$. As time evolves, nodes and edges may appear or disappear. 
In addition, we mask the $\mathcal{G}_t$ via a masking strategy to get the perturbed graph $\mathcal{G}_t^{I}=(\mathcal{V}_t^{I}, \mathcal{E}_t^{I}, \mathcal{X}_t)$ with perturbed adjacency matrix $\mathcal{A}_t^{I}$ and masked graph $\mathcal{G}_t^{B}=(\mathcal{V}_t^{B}, \mathcal{E}_t^{B}, \mathcal{X}_t)$ with masked adjacency matrix $\mathcal{A}_t^{B}$. Specifically, we generate informative subgraph $\mathcal{G}_t^{I}$ using a generative model to serve as the perturbed graph for each snapshot $\mathcal{G}_t$. The complement of the informative subgraph is denoted as the bias subgraph $\mathcal{G}_t^{B}$, which also functions as the masked graph. Note that $\mathcal{G}_t=\mathcal{G}_t^{I} \cup \mathcal{G}_t^{B}$. Essentially, we aim to learn a low-dimensional representation $\mathbf{Z}_t^v \in \mathcal{R}^D$ ($D$ is the node embedding dimension) which captures spatio-temporal patterns for each node $v \in \mathcal{V}_t$ at timestamp $t$ to perform various downstream tasks in dynamic graphs.

\section{Methodology}
In this section, we delve into the specifics of our proposed approach. The overview of the proposed DyGIS is illustrated in Figure \ref{fig2}. Specifically, we leverage an informative subgraph generator to generate an informative subgraph for every snapshot. Meanwhile, a temporal information extraction module is integrated to pass the hidden state in the dynamic graph. At last, we feed the informative subgraph and hidden state extracted from the informative subgraph via the time information extraction module to dynamic graph masked auto-encoder (DGMAE), which yields node representation applied to downstream tasks such as link prediction, and node classification. 

\subsection{Informative Subgraph Generator}
Due to the random masking strategy removing informative subgraphs, potentially leading to the loss of vital information in dynamic graphs, we intuitively consider informative subgraphs as input for DGMAE. This approach harnesses the powerful modeling capabilities of MAE while preventing the loss of essential information in dynamic graphs. 

Particularly, given a dynamic graph $G=\left\{\mathcal{G}_1, \mathcal{G}_2,\cdots, \mathcal{G}_T\right\}$, our idea is that: we design a generative probabilistic model to divide the dynamic graph into informative subgraphs $G^{I}=\left\{\mathcal{G}_1^{I}, \mathcal{G}_2^{I},\cdots, \mathcal{G}_T^{I}\right\}$ and bias subgraphs $G^{B}=\left\{\mathcal{G}_1^{B}, \mathcal{G}_2^{B},\cdots, \mathcal{G}_T^{B}\right\}$. To achieve this objective, we introduce a noisy random graph $\mathcal{G}_t^{R}$ for each snapshot $\mathcal{G}_t$ that shares the same statistical characteristics (\textit{i.e.}, the number of nodes and edges ). Subsequently, we generate informative subgraph $\mathcal{G}_t^{I}$ by maximizing the mutual information between the node embedding $\mathbf{Z}_t$ of $\mathcal{G}_t$ and noisy node embedding $\mathbf{Z}_t^{R}$ of $\mathcal{G}_t^{R}$. By doing so, we aim to reduce informative spatio-temporal information in $\mathbf{Z}_t$, which will result in the poor reconstruction of informative edges. Based on this, we can generate the informative subgraphs.  The core idea of DyGIS is to generate informative subgraphs that guide the evolution of dynamic graphs, which requires the model with generative capabilities. Therefore, we adopt a variational-based framework.

\subsubsection{Inference} With the DyGIS framework, node embedding $\mathbf{Z}_t$ of snapshot $\mathcal{G}_t$ can be sampled by the approximate posterior distribution. Specifically, the inference process can be formalized as follows, 
\begin{equation}q(\mathbf{Z}_{t}|\mathcal{X}_{t}, \mathcal{A}_t, \mathbf{H}_{t-1}) = \prod_{i=1}^{n} q(\mathbf{Z}_{t}^{i}| \mathcal{X}_t, \mathcal{A}_t, \mathbf{H}_{t-1}),\label{eq3}
\end{equation}
\begin{equation}q(\mathbf{Z}_{t}^{i}| \mathcal{X}_t, \mathcal{A}_t, \mathbf{H}_{t-1}) = \mathcal{N}(\mathbf{Z}_{t}^{i}| \boldsymbol{\mu}_{t}^{i}, diag((\boldsymbol{\sigma}^{i}_{t})^{2})),
\end{equation}
where $\mathbf{H}_{t-1}$ is the hidden state of $\mathcal{G}_{t-1}$, $\mathbf{Z}_{t}^{i}$ is the \textit{i}-th of node embedding $\mathbf{Z}_t$, $\boldsymbol{\mu}_{t}^{i}$ and $(\boldsymbol{\sigma}^{i}_{t})^{2}$ are the \textit{i}-th row of mean vetors $\boldsymbol{\mu}_{t}$ and variance vector $(\boldsymbol{\sigma}_{t})^{2}$. For $\mathcal{G}_t^{R}$, We utilize Erdős-Rényi graph model \cite{erdHos1960evolution} to generate a noisy random graph $\mathcal{G}_t^{R}$ with the same number of edges and nodes as $\mathcal{G}_t$. Similar to the inference process of $\mathbf{Z}_t$, noisy node embedding $\mathbf{Z}_t^{R}$ will also be learned through inferring corresponding approximate posterior distribution. More specifically,
\begin{equation}q(\mathbf{Z}_{t}^{R}|\mathcal{X}_{t}, \mathcal{A}_t^{R}, \mathbf{H}_{t-1}) = \prod_{i=1}^{n} q((\mathbf{Z}_{t}^{R})^{i}| \mathcal{X}_t, \mathcal{A}_t^{R}, \mathbf{H}_{t-1}), \label{eq5}
\end{equation} 
\begin{equation}q((\mathbf{Z}_{t}^{R})^{i}| \mathcal{X}_t, \mathcal{A}_t^{R}, \mathbf{H}_{t-1}) = \mathcal{N}((\mathbf{Z}_{t}^{R})^{i}| (\boldsymbol{\mu}_{t}^{R})^{i}, diag(((\boldsymbol{\sigma}_{t}^{R})^{i})^{2})).
\end{equation}
where the hidden states $\mathbf{H}_{t-1}$ used for inferring $\mathbf{Z}_{t}^R$ also originate from $\mathcal{G}_{t-1}$, $\mathcal{X}_t$ remains consistent with $\mathcal{G}_t$. We only alter the topological structure of $\mathcal{G}_t$ to obtain $\mathbf{Z}_t^{R}$. 

More specifically, we employ a two-layer GCN \cite{kipf2016semi} to implement the aforementioned inference process. Simultaneously, the widely used reparameterization trick is applied to replace sample operation, enabling the model back-propagate. 
\begin{equation}
    \boldsymbol{\mu}_{t}  = {\rm GCN}_{\mu}(\mathcal{A}_t, {\rm GCN}(\mathcal{A}_t, [\varphi_{x}(\mathcal{X}_t), \mathbf{H}_{t-1}])),
\end{equation}
\begin{equation}
    \boldsymbol{\sigma}_{t}^{2}  = {\rm GCN}_{\boldsymbol{\sigma}}(\mathcal{A}_t, {\rm GCN}(\mathcal{A}_t, [\varphi_{x}(\mathcal{X}_t), \mathbf{H}_{t-1}])),
\end{equation}
where $\boldsymbol{\mu}_{t}$ and $\boldsymbol{\sigma}_{t}^{2}$ are the mean vectors and variance vectors of $\mathcal{G}_t$. $\varphi_{x}$ is an MLP to further extract features information from $\mathcal{X}_t$ and [] is the concatenation operation. We use a set of shared-parameter encoders to obtain noisy mean vectors $\boldsymbol{\mu}_{t}^{R}$ and variance vectors $(\boldsymbol{\sigma}_{t}^{R})^{2}$ of $\mathcal{G}_t^{R}$. 

\subsubsection{Generation}
After obtaining the node embedding $\mathbf{Z}_t$ of snapshot $\mathcal{G}_t$, we generate the probabilistic graph $\mathcal{G}_t^{Gen}$ with entries between 0 and 1 via $\mathbf{Z}_t$. Specifically, the generation process can be formalized as follows,
\begin{equation}p(\mathcal{A}_{t}^{Gen}|f(\mathbf{Z}_{t})) = \prod_{i=1}^{n} \prod_{j=1}^{n}p(\mathcal{{A}}_{t}^{ij}|f(\mathbf{Z}_{t}^i), f(\mathbf{Z}_{t}^{j})), \label{eq9}
\end{equation}
\begin{equation}p((\mathcal{A}_{t}^{Gen})^{ij}=p |f(\mathbf{Z}_{t}^{i}), f(\mathbf{Z}_{t}^{j})) = \delta(f(\mathbf{Z}_{t}^{i}), f(\mathbf{Z}_{t}^{j})),
\end{equation}
where $f$ is the decoder, we utilize a two-layer MLP as the decoder function. $\mathcal{A}_{t}^{Gen}$ is the adjacency matrix of $\mathcal{G}_t^{Gen}$ and $(\mathcal{A}_{t}^{Gen})^{ij}=p$ denotes the reconstruction probability value of $\mathcal{A}_t^{ij}$ equal to $p$. $\delta$ is the logistic sigmoid function. 

We perturb the spatio-temporal information of the node embedding $\mathbf{Z}_t$ by introducing a noisy random graph $\mathcal{G}_{t}^{R}$, and then use $\mathbf{Z}_t$ to reconstruct $\mathcal{G}_t$. Meanwhile, we aim to maximize the reconstruction probability values for all edges as much as possible via minimized reconstruction loss. Under this constraint, most edges tend to have larger values in $\mathcal{A}_t^{Gen}$. Since $\mathbf{Z}_t$ loses informative spatio-temporal information, the corresponding informative edges can't be reconstructed well compared to the majority of edges. Consequently, the edges in $\mathcal{G}_t^{Gen}$ with lower values are more informative. Based on the above analysis, we first perform the Hadamard product operation on $\mathcal{A}_t$ and $\mathcal{A}_t^{Gen}$. Then we select the edges with the lowest values to construct the informative subgraph $\mathcal{G}_t^{I}$ and collect the complement of  $\mathcal{G}_t^{I}$ as bias subgraph $\mathcal{G}_t^{B}$,  particularly,
\begin{equation}
    \mathcal{E}_{t}^{I} = {\rm TOP}_{r}((1-\mathcal{A}_{t}^{Gen}) \odot \mathcal{A}_{t}) , \label{eq11}
\end{equation}
\begin{equation}
     \mathcal{E}_{t}^{B} = {\rm TOP}_{1-r}(\mathcal{A}_{t}^{Gen} \odot \mathcal{A}_{t}) , \label{eq12}
\end{equation}
where $\mathcal{E}_{t}^{I}$ and $\mathcal{E}_{t}^{B}$ are the edges of $\mathcal{G}_{t}^{I}$ and $\mathcal{G}_{t}^{B}$, respectively. ${\rm TOP}_{r}(\cdot)$ selects the top-\textit{K} edges with $\textit{K}= \textit{r} \times |\mathcal{E}_t|$,  $|\mathcal{E}_t|$ is the number of edges of $\mathcal{G}_t$ and \textit{r} is the informative subgraph ratio. $\odot$ is the element-wise product, \textit{i.e.}, Hadamard product. After obtaining the set of edges, we construct the corresponding subgraph from the set of edges and their associated nodes.

\subsubsection{Learning}
The loss function of the informative subgraph generator consists of two parts. On the one hand, we generate corresponding informative subgraphs based on the performance of graph reconstruction. In the variational framework, the reconstruction loss of dynamic graphs corresponds to the variational lower bound for each snapshot.
\begin{equation}
\begin{split}
\mathcal{L}_{t}^{(1)} &= {\rm KL}(q(\mathbf{Z}_{t}|\mathcal{X}_{t}, \mathcal{A}_t, \mathbf{H}_{t-1})||p(\mathbf{Z}_{t}|\mathbf{H}_{t-1})) \\ 
&- \mathbb{E}_{q(\mathbf{Z}_{t}|\mathcal{X}_{t}, \mathcal{\widetilde{A}}_t, \mathbf{H}_{t-1})}({\rm log}\, p(\mathcal{A}_{t}^{Gen}|f(\mathbf{Z}_{t}))) , \label{16}
\end{split}
\end{equation}
where ${\rm KL}(\cdot||\cdot)$ is the Kullback-Leibler (KL) divergence, $p(\mathbf{Z}_{t}|\mathbf{H}_{t-1})$ is the prior distribution. To enable the model to capture more complex spatio-temporal information, we construct a prior distribution $p(\mathbf{Z}_{t}|\mathbf{H}_{t-1}) = \mathcal{N}(\boldsymbol{\mu}_{t}^{p}(\mathbf{H}_{t-1}), \boldsymbol{\sigma}_{t}^{p}(\mathbf{H}_{t-1})$ by learning the hidden states from previous time steps. Here, $\boldsymbol{\mu}_{t}^{p}(\mathbf{H}_{t-1})$ and $\boldsymbol{\sigma}_{t}^{p}(\mathbf{H}_{t-1})$ are the mean and covariance of the prior distribution learned from hidden state $\mathbf{H}_{t-1}$ via a neural network module. 

On the other hand, By maximizing the mutual information between $\mathbf{Z}_t$ and $\mathbf{Z}_t^{R}$, we aim to enhance the correlation between the two, making $\mathbf{Z}_t$ more inclined towards randomness, thereby perturbing the crucial topological information and temporal dependencies of $\mathbf{Z}_t$. According to Eq.\ref{eq2}, this objective can be achieved by maximizing InfoNCE loss,
\begin{equation}
    \mathcal{L}_{t}^{(2)} = \mathcal{L}^{\mathcal{D}}(\mathbf{Z}_t, \mathbf{Z}_{t}^{R})
\end{equation}
where $\mathcal{L}_t^{(2)}$ is the lower bound of $I(\mathbf{Z}_t;\mathbf{Z}_t^{R})$. And the discriminator $\mathcal{D}$ of the $\mathcal{L}^{D}$ is defined as,
\begin{equation}
    \mathcal{D}(\mathbf{Z}_t, \mathbf{Z}_t^{R}) = sim(\mathbf{Z}_t, \mathbf{Z}_t^{R}) / \tau
\end{equation}
where $sim(\cdot)$ denotes the similarity function, and $\tau$ is the temperature hyper-parameter.

We use the weighted sum of the above two loss functions to guide the model optimization and learn the informative subgraphs. $\lambda$ is used as the trade-off weight hyper-parameter.
\begin{equation}
    \mathcal{L}^{I} = \sum _{t = 1}^{T} \mathcal{L}_{t}^{(1)} -  \lambda \mathcal{L}_{t}^{(2)}. \label{eq16}
\end{equation}

\subsection{Temporal Information Extraction Module}
The temporal information extraction module receives the extracted feature and node embeddings as well as the hidden state from the previous step as the input. Due to the temporal evolution characteristics within dynamic graphs, we adopt  Graph Convolutional Recurrent Networks (GCRN) equipped with gated recurrent units (GRU) and GCN to update spatio-temporal interaction patterns in dynamic graphs. More specifically,
\begin{equation}
 \begin{gathered}
\mathbf{R}_{t}^{g} = \delta_{1}(f_{g}(\mathcal{\widetilde A}_t, [\varphi_{x}(\mathcal{X}_t), \varphi_{z}(\mathbf{Z}_t)]) + f_{g}(\mathcal{\widetilde A}_t, \mathbf{H}_{t-1}))  \\
\mathbf{Z}_{t}^{g} = \delta_{1}(f_{g}(\mathcal{\widetilde A}_t, [\varphi_{x}(\mathcal{X}_t), \varphi_{z}(\mathbf{Z}_t)]) + f_{g}(\mathcal{\widetilde A}_t, \mathbf{H}_{t-1})) \\
\mathbf{\widetilde{H}}_{t} = \delta_{2}(f_{g}(\mathcal{\widetilde A}_t, [\varphi_{x}(\mathcal{X}_t),\varphi_{z}(\mathbf{Z}_t)])+ f_{g}(\mathcal{\widetilde A}_t, \mathbf{R}_{t}^{g} \odot \mathbf{H}_{t-1})) \\
\mathbf{H}_t = \mathbf{Z}_{t}^{g} \odot \mathbf{H}_{t-1}+ (1- \mathbf{Z}_{t}^{g}) \odot \mathbf{\widetilde{H}}_{t},
\end{gathered}   \label{eq17}
\end{equation}

where $\mathbf{R}_{t}^{g}$ is the reset gate that determines which information should be forgotten, $\mathbf{Z}_{t}^{g}$ is the update gate that determines how much of the previous time step's hidden state information should be retained. $f_g$ denotes the GCN. $\delta_1$ and $\delta_2$ represent the sigmoid activation function and the tanh activation function, respectively. $\varphi_z$ plays the same role as $\varphi_x$ to extract features to enhance the expressive power of the model. With the control of the reset gate and update gate, GRU can selectively retain or discard historical information and update the hidden state based on the current input.

\begin{algorithm}[t]
    \caption{Whole Training Algorithm of DyGIS}
    \label{train}
    \renewcommand{\algorithmicrequire}{\textbf{Input:}}
    \renewcommand{\algorithmicensure}{\textbf{Output:}}
    \begin{algorithmic}[1]
        \REQUIRE A dynamic graph $G=\left\{\mathcal{G}_1, \mathcal{G}_2,\cdots, \mathcal{G}_T\right\}$
        \ENSURE Node representation $ \mathbf{Z}^{I}$ for downstream tasks    
        \STATE  Initialize informative subgraphs set $G^I = \varnothing$,  bias subgraphs set $G^R = \varnothing$
        \FOR{ $t=1$ To $T$}
            \STATE Generate noisy random graph $\mathcal{G}_t^{R}$ utilizing E-R graph model
            \STATE Get node embedding $\mathbf{Z}_{t}$ and noisy node embedding  $\mathbf{Z}_{t}^{R}$ by Eq. \ref{eq3} and Eq. \ref{eq5}
            \STATE Get the current hidden state $\mathbf{H}_{t}$ by Eq. \ref{eq17}
            \STATE Generate reconstructed graph $\mathcal{A}_t^{Gen}$ by Eq. \ref{eq9}
            \STATE Generate informative subgraph $\mathcal{G}_t^I$ and bias subgraph $\mathcal{G}_t^B$
            \STATE $G^I = G^I \cup \mathcal{G}_t^I, G^B = G^B \cup \mathcal{G}_t^B$
            \STATE Optimize this process by Eq. \ref{eq16}
        \ENDFOR
      \FOR{ $t=1$ To $T$}
            \STATE Get final node representation $\mathbf{Z}_{t}^{I}$ for downstream tasks by utilizing DGMAE to encode $\mathcal{G}_{t}^{I}$ and $\mathbf{H}_{t-1}^{I}$.
            \STATE Get the current informative hidden state $\mathbf{H}_{t}^{I}$ by Eq. \ref{eq17}
            \STATE Reconstruct the bias subgraph $\mathcal{G}_{t}^{B}$ by utilizing DGMAE to decode
            \STATE Optimize this process by Eq. \ref{eq18}
        \ENDFOR
        \RETURN Node representation $\mathbf{Z}^I = \left\{\mathbf{Z}_1^I, \mathbf{Z}_2^I,\cdots, \mathbf{Z}_T^I\right\}$
    \end{algorithmic} \label{alg_dygis}
\end{algorithm}

\subsection{Dynamic Graph Masked Auto-Encoder}
The informative subgraphs $G^{I}=\left\{\mathcal{G}_1^{I}, \mathcal{G}_2^{I},\cdots, \mathcal{G}_T^{I}\right\}$ and bias subgraphs $G^{B}=\left\{\mathcal{G}_1^{B}, \mathcal{G}_2^{B},\cdots, \mathcal{G}_T^{B}\right\}$ are generated through the informative subgraph generator. We employ the MAE framework to obtain the final node representation to perform various downstream tasks. In particular, like most GMAE paradigms, we utilize the unmasked portion $G^{I}$ to reconstruct the masked portion $G^{B}$. In addition, we still adopt a variational architecture to maintain the consistency of the model.
\subsubsection{Encoder} We feed $\mathcal{G}_t^{I}=(\mathcal{V}_t^{I}, \mathcal{E}_t^{I}, \mathcal{X}_t)$ and informative hidden state $\mathbf{H}_{t-1}^{I}$ extracted with $\mathcal{G}_t^{I}$ via GCRN to the encoder. The output is node representation $\mathbf{Z}_t^{I}$. Similar to the informative subgraph generator, the prior distribution of  DGMAE is also learned from the informative hidden state of the previous time step, \textit{i.e.}, $p(\mathbf{Z}_{t}^{I}|\mathbf{H}_{t-1}^{I}) = \mathcal{N}(\boldsymbol{\mu}_{t}^{p}(\mathbf{H}_{t-1}^{I}), \boldsymbol{\sigma}_{t}^{p}(\mathbf{H}_{t-1}^{I}))$. We enforce the approximate posterior distribution of DGMAE $q(\mathbf{Z}_{t}^{I}|\mathcal{X}_{t}, \mathcal{A}_t^{I}, \mathbf{H}_{t-1}^{I})$ to be close to the prior by minimizing their KL divergence. Additionally, We adopt a two-layer GCN as the encoding function for DGMAE, which is the same as the one in the informative subgraph generator but without sharing parameters. 
\subsubsection{Decoder} We aim to explore MAE on dynamic graphs, specifically using the unmasked portions to reconstruct the masked portions to learn effective node embedding. Therefore, in the decoder, node representation $\mathbf{Z}_t^{I}$ is the input, and the output is the reconstructed bias subgraph $\mathcal{G}_t^{B}$. We employ a single-layer GCN and inner product as the decoder function. We optimize the model by minimizing both the reconstruction error and the KL divergence.
\begin{equation}
\begin{split}
    \mathcal{L}^{M} &= \sum_{t=1}^{T}[-\frac{1}{\left|\mathcal{E}_{t}^{B}\right|}\sum_{(v, u) \in \mathcal{E}_{t}^{B}} \log \frac{\exp (g(v, u))}{\sum_{w \in \mathcal{V}_{t}} \exp (g(v, w))} \\
    &+ {\rm KL}(q(\mathbf{Z}_{t}^{I}|\mathcal{X}_{t}, \mathcal{A}_t^{I}, \mathbf{H}_{t-1}^{I})||p(\mathbf{Z}_{t}^{I}|\mathbf{H}_{t-1}^{I}))] \label{eq18}
\end{split}
\end{equation}
where $g(\cdot)$ is the link probability estimator between two nodes. 
$\mathcal{\overline{G}}_t $.  Algorithm\ref{alg_dygis} presents the overall process of DyGIS.


\section{Experiments}
In this section, we conduct numerous experiments to validate the superiority of DyGIS\footnote{Code is available at https://github.com/KeepMovingXX/DyGIS} across different tasks. Additionally, we perform ablation experiments and parameter sensitivity analysis to verify the effectiveness of each component of the model. Lastly, we illustrate how DyGIS generates what kind of subgraph serves as the informative subgraph through a case study. 

\begin{table}[t]
\caption{Characteristics of datasets.}
\label{tab1}
\resizebox{\linewidth}{!}{
\begin{tabular}{cccccc}
\toprule
\textbf{Dataset}                       & \textbf{Nodes} & \textbf{Edges} & \textbf{Snapshots} & \textbf{Test $l$} & \textbf{Class}  \\ \midrule
\textbf{Enron}                         & 184            & 4,784            & 11                 & 3                 & -                         \\ 
\textbf{DBLP}                          & 315            & 5,104            & 10                 & 3                 & -                       \\ 
\textbf{FB}                            & 663            & 23,394           & 9                  & 3                 & -                         \\ 
\textbf{Email}                         & 2,029           & 39,264          & 29                 & 3                & -                         \\ 
\textbf{Socwiki}                      & 8,298           & 106,043        & 12                 & 3                 & -                \\ 
\multicolumn{1}{c}{\textbf{Iadublin}} & 10,973          & 415,913        & 8                  & 3                 & -                \\ \midrule
\textbf{HS12}                          & 180            & 8,608            & 8                  & 3                 & 7                         \\ 
\textbf{Primary}                       & 242            & 35,734           & 6                  & 3                 & 13                        \\ 
\textbf{HighSchool}                    & 327            & 26,870         & 9                  & 3                 & 7                         \\ 
\textbf{CellPhone}                    & 400            & 10,248           & 10                 & 3                 & 20                     \\ 
\textbf{Cora}                          & 2,708           & 16,279           & 5                  & 1                 & 7               \\ \bottomrule
\end{tabular}}

\end{table}

\subsection{Datasets and Baselines}
We conduct experiments on eleven datasets with varying scales to evaluate our model and baselines. The characteristics of datasets are shown in Table \ref{tab1}. DyGIS is a DyGNN method based on masked autoencoders. Therefore, we first select advanced DyGNN methods as baselines to show the superiority of DyGIS. These methods include VGRNN \cite{hajiramezanali2019variational}, GRUGCN \cite{seo2018structured}, EvolveGCN \cite{pareja2020evolvegcn}, DySAT \cite{sankar2020dysat}, HTGN \cite{yang2021discrete}, DGCN \cite{gao2023novel}, HGWaveNet \cite{bai2023hgwavenet}, DyTed \cite{zhang2023dyted}. Among the DyGNN methods, DGCN and DyTed are the contrastive SSL DyGNN methods. since our datasets include dynamic graphs without features, GraphMAE \cite{hou2022graphmae} and GraphMAE2 \cite{hou2023graphmae2} focus on masked features and are not applicable to these datasets. Therefore, we select two state-of-the-art masked graph structure methods (S2GAE \cite{li2023s} and MaskGAE \cite{tan2023s2gae}) as baselines in our paper. In addition, we compare with GAE and VGAE \cite{kipf2016variational} to demonstrate how temporal dependencies and the MAE framework can improve performance. For all baselines, we follow the setting described in their original papers.

\begin{table*}[t]
\caption{AUC and AP scores of link detection. The best are bolded and the second best are underlined.}
\label{table_dection}
\resizebox{\textwidth}{!}{
\begin{tabular}{c|cc|cc|cc|cc|cc|cc}
\toprule
Dataset   & \multicolumn{2}{c|}{Enron}                                     & \multicolumn{2}{c|}{DBLP}                                      & \multicolumn{2}{c|}{FB}                                        & \multicolumn{2}{c|}{Email}                                     & \multicolumn{2}{c|}{Socwiki}                                   & \multicolumn{2}{c}{Iadublin}                                   \\
Metric    & AUC                                      & AP                  & AUC                                      & AP                  & AUC                                      & AP                  & AUC                                      & AP                  & AUC                                      & AP                  & AUC                                      & AP                  \\ \midrule
GAE       & \multicolumn{1}{c|}{88.94±2.73}          & 88.03±2.28          & \multicolumn{1}{c|}{76.43±2.39}          & 76.99±2.42          & \multicolumn{1}{c|}{80.60±1.46}          & 78.96±1.55          & \multicolumn{1}{c|}{88.65±2.12}          & 90.31±2.01          & \multicolumn{1}{c|}{77.26±0.22}          & 77.34±0.22          & \multicolumn{1}{c|}{82.62±1.42}          & 81.57±2.45          \\
VGAE      & \multicolumn{1}{c|}{90.43±3.12}          & 89.63±3.29          & \multicolumn{1}{c|}{77.55±3.13}          & 79.64±2.52          & \multicolumn{1}{c|}{83.26±1.14}          & 82.73±1.54          & \multicolumn{1}{c|}{89.43±3.17}          & 91.22±1.43          & \multicolumn{1}{c|}{77.85±3.24}          & 80.62±3.87          & \multicolumn{1}{c|}{83.25±1.68}          & 77.25±2.71          \\ \midrule
VGRNN     & \multicolumn{1}{c|}{94.32±0.52}          & 95.28±0.63          & \multicolumn{1}{c|}{87.39±2.95}    & 88.81±2.51          & \multicolumn{1}{c|}{87.78±0.93}          & 87.35±1.11          & \multicolumn{1}{c|}{88.42±2.05}          & 91.54±1.42          & \multicolumn{1}{c|}{85.10±2.13}          & \underline{90.89±0.95}    & \multicolumn{1}{c|}{87.31±1.91}          & 79.75±2.93          \\
GRUGCN    & \multicolumn{1}{c|}{94.05±1.05}          & 94.69±0.63          & \multicolumn{1}{c|}{77.50±2.33}          & 81.71±1.85          & \multicolumn{1}{c|}{84.16±1.36}          & 83.13±1.37          & \multicolumn{1}{c|}{80.50±2.13}          & 87.82±1.55          & \multicolumn{1}{c|}{88.07±3.58}    & 87.70±5.28          & \multicolumn{1}{c|}{80.22±3.78}          & 75.79±5.79          \\
EvolveGCN & \multicolumn{1}{c|}{75.39±3.96}          & 72.99±1.89          & \multicolumn{1}{c|}{69.37±2.94}          & 70.14±3.45          & \multicolumn{1}{c|}{72.40±1.55}          & 69.81±1.79          & \multicolumn{1}{c|}{74.17±3.71}          & 77.22±4.09          & \multicolumn{1}{c|}{74.53±4.90}          & 73.03±6.09          & \multicolumn{1}{c|}{65.60±2.74}          & 62.63±1.99          \\
DySAT     & \multicolumn{1}{c|}{94.41±1.71}          & 93.79±1.64          & \multicolumn{1}{c|}{77.43±2.18}          & 81.23±1.85          & \multicolumn{1}{c|}{86.39±0.69}          & 87.27±0.51          & \multicolumn{1}{c|}{80.02±2.57}          & 88.54±1.42          & \multicolumn{1}{c|}{84.13±1.09}          & 83.96±1.07          & \multicolumn{1}{c|}{70.56±3.32}          & 58.32±2.78          \\
HTGN      & \multicolumn{1}{c|}{94.61±1.65}          & 95.40±1.43          & \multicolumn{1}{c|}{83.18±1.60}          & 86.37±1.46          & \multicolumn{1}{c|}{84.49±1.30}          & 83.09±1.74          & \multicolumn{1}{c|}{\textbf{93.98±0.86}} & \textbf{95.11±0.75} & \multicolumn{1}{c|}{81.82±1.28}          & 80.76±1.47          & \multicolumn{1}{c|}{80.18±1.73}          & 82.81±1.70          \\
HGWaveNet & \multicolumn{1}{c|}{\underline{95.90±1.24}}    & \underline{95.76±1.33}    & \multicolumn{1}{c|}{86.14±0.97}          & 89.42±1.29    & \multicolumn{1}{c|}{\underline{87.97±0.91}}    & 85.38±1.64          & \multicolumn{1}{c|}{91.02±0.51}          & 93.28±0.87          & \multicolumn{1}{c|}{84.33±1.20}          & 79.54±1.41          & \multicolumn{1}{c|}{\underline{90.14±0.80}}    & 86.42±0.91          \\ \midrule
DGCN      & \multicolumn{1}{c|}{86.98±2.61}          & 84.45±1.54          & \multicolumn{1}{c|}{71.91±1.78}          & 73.47±1.25          & \multicolumn{1}{c|}{75.52±1.28}          & 74.91±2.11          & \multicolumn{1}{c|}{90.44±0.23}          & 90.70±0.35          & \multicolumn{1}{c|}{86.09±0.72}          & 88.79±0.81          & \multicolumn{1}{c|}{88.99±1.05}          & \underline{88.22±0.85}    \\
DyTed          & \multicolumn{1}{c|}{91.32±1.29} & 91.90±1.72 & \multicolumn{1}{c|}{\underline{88.84±1.16}} & \underline{90.10±0.74}   & \multicolumn{1}{c|}{85.53±1.27} & 87.09±1.17 & \multicolumn{1}{c|}{84.68±1.59} & 81.98±1.26 & \multicolumn{1}{c|}{\underline{90.25±0.37}} & 89.37±0.52 & \multicolumn{1}{c|}{73.56±0.39} & 74.55±0.19 \\
\midrule
S2GAE     & \multicolumn{1}{c|}{93.88±1.24}          & 94.34±1.40          & \multicolumn{1}{c|}{76.86±3.29}          & 73.20±4.80          & \multicolumn{1}{c|}{82.20±1.48}          & 80.31±0.98          & \multicolumn{1}{c|}{90.61±1.47}          & 91.37±0.82          & \multicolumn{1}{c|}{77.98±0.53}          & 83.65±0.32          & \multicolumn{1}{c|}{76.55±2.75}          & 82.65±1.73          \\
MaskGAE   & \multicolumn{1}{c|}{94.72±0.27}          & 94.07±0.53          & \multicolumn{1}{c|}{81.18±0.49}          & 84.73±0.42          & \multicolumn{1}{c|}{87.88±0.48}          & \underline{88.74±0.64}    & \multicolumn{1}{c|}{90.38±0.90}          & 91.48±0.82          & \multicolumn{1}{c|}{85.81±0.06}          & 84.88±0.05          & \multicolumn{1}{c|}{85.33±0.43}          & 86.75±0.64          \\ \midrule
DyGIS      & \multicolumn{1}{c|}{\textbf{97.25±0.78}} & \textbf{97.28±0.73} & \multicolumn{1}{c|}{\textbf{92.08±1.55}} & \textbf{92.76±1.38} & \multicolumn{1}{c|}{\textbf{92.97±0.44}} & \textbf{92.84±0.63} & \multicolumn{1}{c|}{\underline{92.38±0.64}}    & \underline{94.17±0.67}    & \multicolumn{1}{c|}{\textbf{93.01±0.42}} & \textbf{93.91±0.28} & \multicolumn{1}{c|}{\textbf{92.14±1.65}} & \textbf{92.24±1.50} \\ \bottomrule
\end{tabular}}
\end{table*}

\begin{table*}[t]
\caption{AUC and AP scores of link prediction. The best are bolded and the second best are underlined.}
\label{table_prediction}
\resizebox{\textwidth}{!}{
\begin{tabular}{c|cc|cc|cc|cc|cc|cc}
\toprule
Dataset   & \multicolumn{2}{c|}{Enron}                                     & \multicolumn{2}{c|}{DBLP}                                      & \multicolumn{2}{c|}{FB}                                        & \multicolumn{2}{c|}{Email}                                     & \multicolumn{2}{c|}{Socwiki}                                   & \multicolumn{2}{c}{Iadublin}                                   \\
Metric    & AUC                                      & AP                  & AUC                                      & AP                  & AUC                                      & AP                  & AUC                                      & AP                  & AUC                                      & AP                  & AUC                                      & AP                  \\ \midrule
GAE       & \multicolumn{1}{c|}{92.55±0.76}          & 93.64±0.51          & \multicolumn{1}{c|}{84.71±0.73}          & 87.78±0.42          & \multicolumn{1}{c|}{89.47±0.65}          & 88.93±0.80          & \multicolumn{1}{c|}{81.34±0.18}          & 89.37±0.12          & \multicolumn{1}{c|}{55.81±0.86}          & 57.60±0.58          & \multicolumn{1}{c|}{58.08±0.58}          & 59.05±0.58          \\
VGAE      & \multicolumn{1}{c|}{92.46±0.49}          & 93.64±0.30          & \multicolumn{1}{c|}{85.51±0.44}          & 88.45±0.17          & \multicolumn{1}{c|}{88.91±0.22}          & 88.16±0.29          & \multicolumn{1}{c|}{82.58±0.36}          & 89.95±0.19          & \multicolumn{1}{c|}{58.34±0.51}          & 59.69±0.43          & \multicolumn{1}{c|}{63.61±1.86}          & 57.12±1.70          \\ \midrule
VGRNN     & \multicolumn{1}{c|}{93.31±0.95}          & 93.39±0.90          & \multicolumn{1}{c|}{85.38±0.76}          & 88.07±0.59          & \multicolumn{1}{c|}{89.27±0.56}          & 90.12±0.61          & \multicolumn{1}{c|}{93.89±1.58}          & \underline{95.31±1.05}    & \multicolumn{1}{c|}{60.16±3.07}          & 65.16±1.87          & \multicolumn{1}{c|}{68.83±5.31}          & 69.96±4.71          \\
GRUGCN    & \multicolumn{1}{c|}{92.65±0.88}          & 93.23±0.69          & \multicolumn{1}{c|}{83.67±0.97}          & 85.76±0.94          & \multicolumn{1}{c|}{79.27±0.12}          & 76.86±1.14          & \multicolumn{1}{c|}{80.09±0.38}          & 88.05±0.18          & \multicolumn{1}{c|}{60.63±2.27}          & 63.50±2.09          & \multicolumn{1}{c|}{69.15±2.14}          & 56.80±1.43          \\
EvolveGCN & \multicolumn{1}{c|}{91.37±0.54}          & 92.84±0.41          & \multicolumn{1}{c|}{83.65±0.29}          & 87.23±0.29          & \multicolumn{1}{c|}{85.71±0.51}          & 86.47±0.78          & \multicolumn{1}{c|}{80.86±2.23}          & 86.89±1.90          & \multicolumn{1}{c|}{71.19±6.92}          & 72.53±4.97          & \multicolumn{1}{c|}{\underline{78.37±2.00}}    & 65.09±2.12          \\
DySAT     & \multicolumn{1}{c|}{94.23±0.44}          & 95.05±0.36          & \multicolumn{1}{c|}{86.28±0.49}          & 89.01±0.32          & \multicolumn{1}{c|}{89.76±0.19}          & 89.29±0.21          & \multicolumn{1}{c|}{86.69±0.28}          & 92.22±0.16          & \multicolumn{1}{c|}{57.22±0.83}          & 58.51±0.56          & \multicolumn{1}{c|}{66.21±2.37}          & 54.73±1.76          \\
HTGN      & \multicolumn{1}{c|}{94.42±0.35}          & 94.64±0.29          & \multicolumn{1}{c|}{88.57±0.47}          & 90.98±0.31          & \multicolumn{1}{c|}{87.10±0.23}          & 87.02±0.72          & \multicolumn{1}{c|}{\underline{94.08±0.28}}    & 94.90±0.23          & \multicolumn{1}{c|}{75.54±0.77}          & 76.62±0.70          & \multicolumn{1}{c|}{71.45±2.78}          & 65.92±2.44          \\
HGWaveNet & \multicolumn{1}{c|}{\underline{95.32±0.19}}    & \underline{95.41±0.26}    & \multicolumn{1}{c|}{\underline{89.51±0.18}}    & \underline{91.38±0.18}    & \multicolumn{1}{c|}{87.72±0.33}          & 86.05±0.52          & \multicolumn{1}{c|}{92.67±0.36}          & 93.84±0.29          & \multicolumn{1}{c|}{71.56±2.17}          & 73.01±2.21          & \multicolumn{1}{c|}{66.26±2.39}          & 62.22±2.75          \\ \midrule
DGCN      & \multicolumn{1}{c|}{83.36±0.62}          & 80.31±0.83          & \multicolumn{1}{c|}{75.06±1.14}          & 73.56±1.08          & \multicolumn{1}{c|}{71.77±1.15}          & 71.92±2.21          & \multicolumn{1}{c|}{93.77±0.69}          & 92.47±0.76          & \multicolumn{1}{c|}{\underline{86.25±0.55}}    & \underline{80.18±0.69}    & \multicolumn{1}{c|}{74.22±0.32}          & \underline{76.60±0.51}    \\
DyTed          & \multicolumn{1}{c|}{92.12±0.88} & 92.54±1.10 & \multicolumn{1}{c|}{86.24±0.63} & 88.10±0.93 & \multicolumn{1}{c|}{88.35±1.73} & 86.13±1.35 & \multicolumn{1}{c|}{84.27±1.49} & 86.02±1.13 & \multicolumn{1}{c|}{76.66±1.08} & 78.99±0.94 & \multicolumn{1}{c|}{77.31±0.67} & 76.48±0.87 \\
\midrule
S2GAE     & \multicolumn{1}{c|}{93.57±0.13}          & 93.42±0.22          & \multicolumn{1}{c|}{88.46±0.29}          & 90.52±0.15          & \multicolumn{1}{c|}{90.90±0.08}          & 90.61±0.11          & \multicolumn{1}{c|}{93.23±0.31}          & 94.22±0.19          & \multicolumn{1}{c|}{59.02±0.43}          & 59.18±0.32          & \multicolumn{1}{c|}{62.60±0.59}          & 62.17±0.56          \\
MaskGAE   & \multicolumn{1}{c|}{94.06±0.19}          & 94.91±0.19          & \multicolumn{1}{c|}{87.98±0.20}          & 89.96±0.19          & \multicolumn{1}{c|}{\underline{90.33±0.30}}    & \underline{89.98±0.32}    & \multicolumn{1}{c|}{93.04±0.07}          & 94.30±0.03          & \multicolumn{1}{c|}{68.09±0.39}          & 69.77±1.21          & \multicolumn{1}{c|}{66.68±0.41}          & 67.14±0.13          \\ \midrule
DyGIS      & \multicolumn{1}{c|}{\textbf{95.90±0.29}} & \textbf{95.48±0.37} & \multicolumn{1}{c|}{\textbf{95.55±0.58}} & \textbf{95.37±0.60} & \multicolumn{1}{c|}{\textbf{93.61±0.17}} & \textbf{92.30±0.19} & \multicolumn{1}{c|}{\textbf{97.48±0.54}} & \textbf{97.87±0.42} & \multicolumn{1}{c|}{\textbf{92.40±0.29}} & \textbf{91.85±0.27} & \multicolumn{1}{c|}{\textbf{82.59±0.62}} & \textbf{87.20±0.60} \\ \bottomrule
\end{tabular}}
\end{table*}

\subsection{Evaluation Tasks and Metric} We first evaluate our model and baselines on three widely studied link prediction tasks in dynamic graphs. Specifically, given partially observed snapshots of a dynamic graph $G = \left\{\mathcal{G}_1, \mathcal{G}_2,...\mathcal{G}_t\right\}$. Three different tasks are defined as follows: 1) link detection, \textit{i.e.}, detect unobserved edges with partial edges observed in $\mathcal{G}_t$. 2) link prediction, \textit{i.e.}, predict edges in $\mathcal{G}_{t+1}$. 3) new link prediction, \textit{i.e.}, predict edges in $\mathcal{G}_{t+1}$ that don't exist in $\mathcal{G}_t$. We adopt average precision (AP) and area under the ROC curve (AUC) scores as the metric of three different link prediction tasks. Additionally, to further validate the learning ability of our model, we conduct node classification experiments on five dynamic datasets with ground truth. We use commonly used accuracy (ACC) as the evaluation metric for node classification. Reported experimental results are the mean and standard deviation of 10 runs to avoid random errors.

\subsection{Implement Details} Following the same setting in VGRNN \cite{hajiramezanali2019variational}, we randomly choose 5\% and 10\% of edges at each snapshot as validation and test sets on link detection task. The last $l$ snapshots are split into test sets to verify the performance of our model. For datasets without features, we employ one-hot encoding as features for small-scale datasets. For large-scale datasets (Socwiki, Iadublin), we follow the setting of HTGN \cite{yang2021discrete} and use learnable matrices as features. Note that we only use the perturbed graph (\textit{i.e.}, informative subgraph) for training, while the evaluation is conducted using the original dynamic graph.
The number of epochs is fixed at 100 and 1000 for the informative subgraph generator and dynamic masked auto-encoder, respectively. The embedding dimension $D$ is 32. We set informative subgraph ratio $r$, trade-off weight $\lambda$, and temperature hyper-parameter $\tau$ to 0.1, 0.5, and 0.7, respectively. For three different link prediction tasks, the learning rates are set to 1e-3 and 5e-3 corresponding to different datasets. For node classification tasks, the learning rate is set to 2e-2 and the fine-tuning epochs for the linear classifier are fixed at 300.




\subsection{Experiment Results}
\subsubsection{Link Detection}
The results of link detection are shown in Table \ref{table_dection}. DyGIS outperforms all baselines in terms of both AUC and AP in the majority of cases. Improvements made by DyGIS compared with all DyGNN methods show that mask modeling based on generating informative subgraphs enables the model to capture more meaningful spatio-temporal information into latent node representation. Both S2GAE and MaskGAE utilize a random masking strategy. Comparisons between DyGIS and S2GAE, MaskGAE show that our mask strategy of generating informative subgraphs is more effective than random masking. In most cases, DyGNN methods are generally superior to GAE/VGAE, indicating that capturing temporal dependencies is essential for dynamic graphs. Therefore, when applying MAE to dynamic graphs, it is essential to consider the temporal dependencies in dynamic graphs. Additionally, comparing S2GAE and MaskGAE with GAE/VGAE also suggests that modeling dynamic graphs through masking can capture more meaningful information in most datasets.

\subsubsection{Link Prediction} Table \ref{table_prediction} summarizes the results for link prediction in different datasets. DyGIS outperforms all baselines in terms of AUC and AP across all datasets. It can be observed that the superiority of our model is more significant on large datasets than small ones compared with baselines, especially the two MAE methods. This is because a dynamic graph with larger scales often exhibits stronger temporal dependencies. S2GAE and MaskGAE adopting random masking strategy compromise the informative spatio-temporal information in dynamic graphs, whereas our approach captures informative spatio-temporal information by generating informative subgraphs. Consequently, our proposed DyGIS ensures the integrity of crucial spatio-temporal information in dynamic graphs with complex temporal dependencies.
\begin{table*}[htbp] 
\caption{AUC and AP scores of new link prediction. The best are bolded and the second best are underlined.}
\label{table_new_prediction}
\resizebox{\textwidth}{!}{
\begin{tabular}{c|cc|cc|cc|cc|cc|cc}
\toprule
Dataset   & \multicolumn{2}{c|}{Enron}                                     & \multicolumn{2}{c|}{DBLP}                                      & \multicolumn{2}{c|}{FB}                                          & \multicolumn{2}{c|}{Email}                                     & \multicolumn{2}{c|}{Socwiki}                                   & \multicolumn{2}{c}{Iadublin}                                   \\
Metric    & AUC                                      & AP                  & AUC                                      & AP                  & AUC                                        & AP                  & AUC                                      & AP                  & AUC                                      & AP                  & AUC                                      & AP                  \\ \midrule
GAE       & \multicolumn{1}{c|}{87.57±1.07}          & 87.99±0.86          & \multicolumn{1}{c|}{78.11±1.26}          & 82.15±0.97          & \multicolumn{1}{c|}{88.55±0.53}            & 87.58±0.77          & \multicolumn{1}{c|}{75.73±0.32}          & 85.68±0.26          & \multicolumn{1}{c|}{54.51±0.66}          & 55.92±0.51          & \multicolumn{1}{c|}{56.51±0.51}          & 57.18±0.59          \\
VGAE      & \multicolumn{1}{c|}{87.30±0.82}          & 87.66±0.73          & \multicolumn{1}{c|}{78.81±1.05}          & 82.98±0.50          & \multicolumn{1}{c|}{88.64±0.18}            & 87.59±0.24          & \multicolumn{1}{c|}{77.35±0.57}          & 86.50±0.32          & \multicolumn{1}{c|}{57.02±0.34}          & 57.93±0.27          & \multicolumn{1}{c|}{63.32±1.25}          & 56.97±5.54          \\ \midrule
VGRNN     & \multicolumn{1}{c|}{87.25±1.61}          & 86.33±1.94          & \multicolumn{1}{c|}{76.94±0.24}          & 78.24±0.54          & \multicolumn{1}{c|}{86.39±0.51}            & 85.45±0.43          & \multicolumn{1}{c|}{\underline{92.77±1.23}}    & \underline{94.04±0.80}    & \multicolumn{1}{c|}{62.10±2.92}          & 66.05±1.88          & \multicolumn{1}{c|}{69.31±8.64}          & 70.22±9.08          \\
GRUGCN    & \multicolumn{1}{c|}{86.72±1.52}          & 86.48±1.52          & \multicolumn{1}{c|}{77.53±1.99}          & 79.76±1.59          & \multicolumn{1}{c|}{79.16±0.18}            & 76.10±0.27          & \multicolumn{1}{c|}{73.89±0.42}          & 83.77±0.14          & \multicolumn{1}{c|}{58.84±2.61}          & 61.65±2.46          & \multicolumn{1}{c|}{68.34±2.28}          & 56.16±1.55          \\
EvolveGCN & \multicolumn{1}{c|}{84.79±0.68}          & 85.82±0.53          & \multicolumn{1}{c|}{73.68±0.62}          & 78.04±0.54          & \multicolumn{1}{c|}{82.21±0.83}            & 82.12±0.63          & \multicolumn{1}{c|}{74.50±3.10}          & 81.99±2.15          & \multicolumn{1}{c|}{72.03±6.52}          & 73.24±4.95          & \multicolumn{1}{c|}{\underline{78.21±1.88}}    & 64.94±2.01          \\
DySAT     & \multicolumn{1}{c|}{89.70±0.58}          & 89.52±0.78          & \multicolumn{1}{c|}{79.23±0.84}          & 82.83±0.67          & \multicolumn{1}{c|}{\underline{88.84±0.17}}      & 87.67±0.14          & \multicolumn{1}{c|}{82.30±0.54}          & 89.26±0.29          & \multicolumn{1}{c|}{56.34±0.81}          & 57.49±0.62          & \multicolumn{1}{c|}{65.77±2.37}          & 54.40±1.72          \\
HTGN      & \multicolumn{1}{c|}{90.71±0.36}          & 89.78±0.35          & \multicolumn{1}{c|}{82.81±0.91}          & \underline{85.41±0.73}    & \multicolumn{1}{c|}{84.68±0.23}            & 83.80±0.51          & \multicolumn{1}{c|}{91.38±0.45}          & 92.08±0.44          & \multicolumn{1}{c|}{76.62±0.67}          & 77.81±0.62          & \multicolumn{1}{c|}{70.78±2.96}          & 65.33±2.62          \\
HGWaveNet & \multicolumn{1}{c|}{\underline{91.41±0.40}}    & 90.15±0.31          & \multicolumn{1}{c|}{\underline{83.26±0.22}}    & 84.73±0.21          & \multicolumn{1}{c|}{85.83±0.59}            & 83.79±0.75          & \multicolumn{1}{c|}{89.72±0.34}          & 91.19±0.31          & \multicolumn{1}{c|}{72.64±1.82}          & 74.17±2.10          & \multicolumn{1}{c|}{65.88±2.39}          & 61.88±2.65          \\ \midrule
DGCN      & \multicolumn{1}{c|}{81.25±1.68}          & 77.92±1.12          & \multicolumn{1}{c|}{74.15±1.55}          & 75.31±1.41          & \multicolumn{1}{c|}{72.36±0.68}            & 70.30±1.23          & \multicolumn{1}{c|}{90.33±0.84}          & 90.23±0.71          & \multicolumn{1}{c|}{\underline{86.14±0.48}}    & \underline{80.50±0.41}    & \multicolumn{1}{c|}{74.12±0.58}          &74.53±0.60    \\
DyTed          & \multicolumn{1}{c|}{84.68±1.28} & 85.61±1.15 & \multicolumn{1}{c|}{80.50±0.78} & 81.63±0.65   & \multicolumn{1}{c|}{87.62±2.83} & 83.28±3.25 & \multicolumn{1}{c|}{85.27±3.03} & 86.65±2.95 & \multicolumn{1}{c|}{76.34±1.26} & 78.58±0.91 & \multicolumn{1}{c|}{77.53±0.80} & \underline{76.61±0.92} \\
\midrule
S2GAE     & \multicolumn{1}{c|}{89.99±0.23}          & 88.94±0.30          & \multicolumn{1}{c|}{82.61±0.57}          & 84.33±0.40          & \multicolumn{1}{c|}{88.71±0.10}            & 88.23±0.21          & \multicolumn{1}{c|}{91.87±0.42}          & 93.07±0.15          & \multicolumn{1}{c|}{59.09±0.36}          & 59.35±0.21          & \multicolumn{1}{c|}{59.60±0.27}          & 59.71±0.18          \\
MaskGAE   & \multicolumn{1}{c|}{91.28±0.20}          & \underline{90.22±0.29}    & \multicolumn{1}{c|}{83.05±0.38}          & 84.17±0.29          & \multicolumn{1}{c|}{88.17±0.26}            & \underline{88.55±0.29}    & \multicolumn{1}{c|}{91.84±0.11}          & 93.38±0.13          & \multicolumn{1}{c|}{65.96±0.11}          & 67.94±0.13          & \multicolumn{1}{c|}{58.95±2.33}          & 63.07±0.69          \\ \midrule
DyGIS      & \multicolumn{1}{c|}{\textbf{92.57±0.69}} & \textbf{90.86±0.63} & \multicolumn{1}{c|}{\textbf{93.09±0.32}} & \textbf{92.31±0.38} & \multicolumn{1}{c|}{\textbf{92.53±0.43}} & \textbf{90.76±0.32} & \multicolumn{1}{c|}{\textbf{96.96±0.52}} & \textbf{97.34±0.45} & \multicolumn{1}{c|}{\textbf{92.43±0.35}} & \textbf{91.92±0.39} & \multicolumn{1}{c|}{\textbf{82.42±0.62}} & \textbf{87.05±0.61} \\ \bottomrule
\end{tabular}}
\end{table*}

\begin{figure*}[t]
\centering
\includegraphics[width=1.0\textwidth]{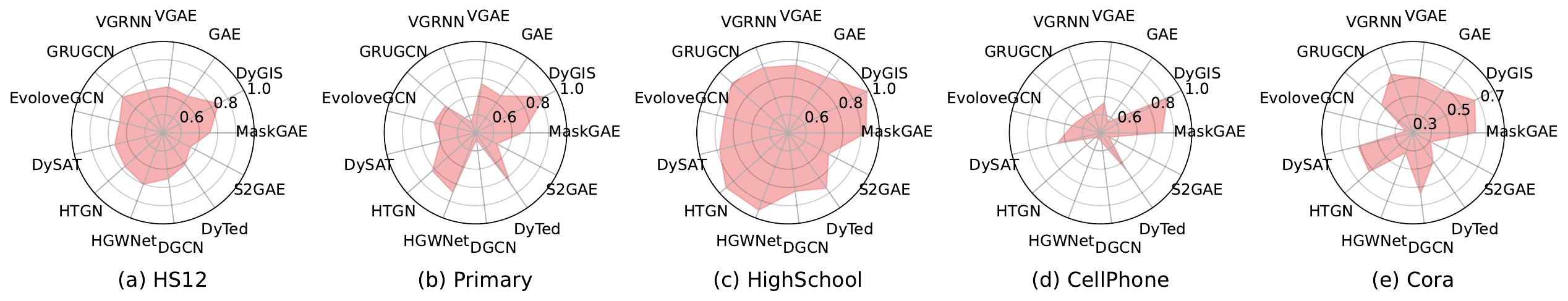} 
\caption{The ACC values of node classification.}
\label{figure_node}
\end{figure*}

\subsubsection{New Link Prediction} This task aims at predicting new edges that are emerging in the next snapshot and evaluating the model’s inductive ability, which is more challenging. Experimental results for the new link prediction task on various datasets are shown in Table \ref{table_new_prediction}. We can conclude similar to link prediction. DyGIS outperforms all baselines on all datasets, once again demonstrating the superiority of our model. In addition, it can be found that the performance of each method degrades to varying degrees compared to the corresponding link prediction task, especially on large-scale datasets, while our model produces more consistent results. This indicates that DyGIS captures the evolving patterns of dynamic graphs through informative subgraphs, achieving better performance in more challenging new link prediction.

\subsubsection{Node Classification} We perform node classification experiments to provide additional validation of the model's learning ability and show results in Figure \ref{figure_node}. We first train DyGIS to obtain node representations, followed by fine-tuning using a linear classifier to derive the experimental results. From Figure \ref{figure_node}, it can be found that our model consistently achieves optimal performance on five datasets, which validates the superiority of DyGIS from a different perspective. Specifically, DyGIS maintains superior learning ability in node classification and outperforms all baselines by a large margin in most cases, which indicates that the node representations learned by DyGIS are more discriminative and effective. 
This serves as additional evidence of the excellent learning ability of our model.
\begin{figure*}[htbp]
\centering
\subfloat[DBLP]{
\centering
\includegraphics[width=0.24\linewidth]{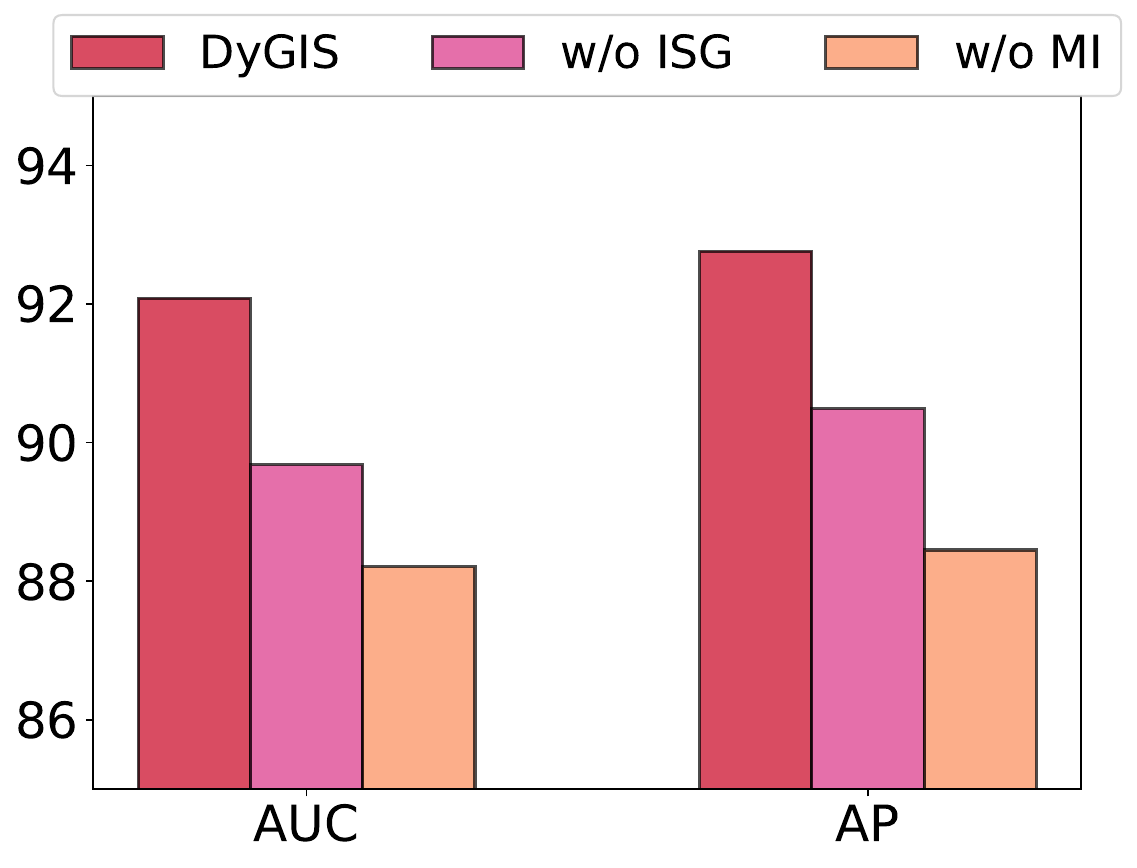}
}%
\subfloat[FB]{
\centering
\includegraphics[width=0.23\linewidth]{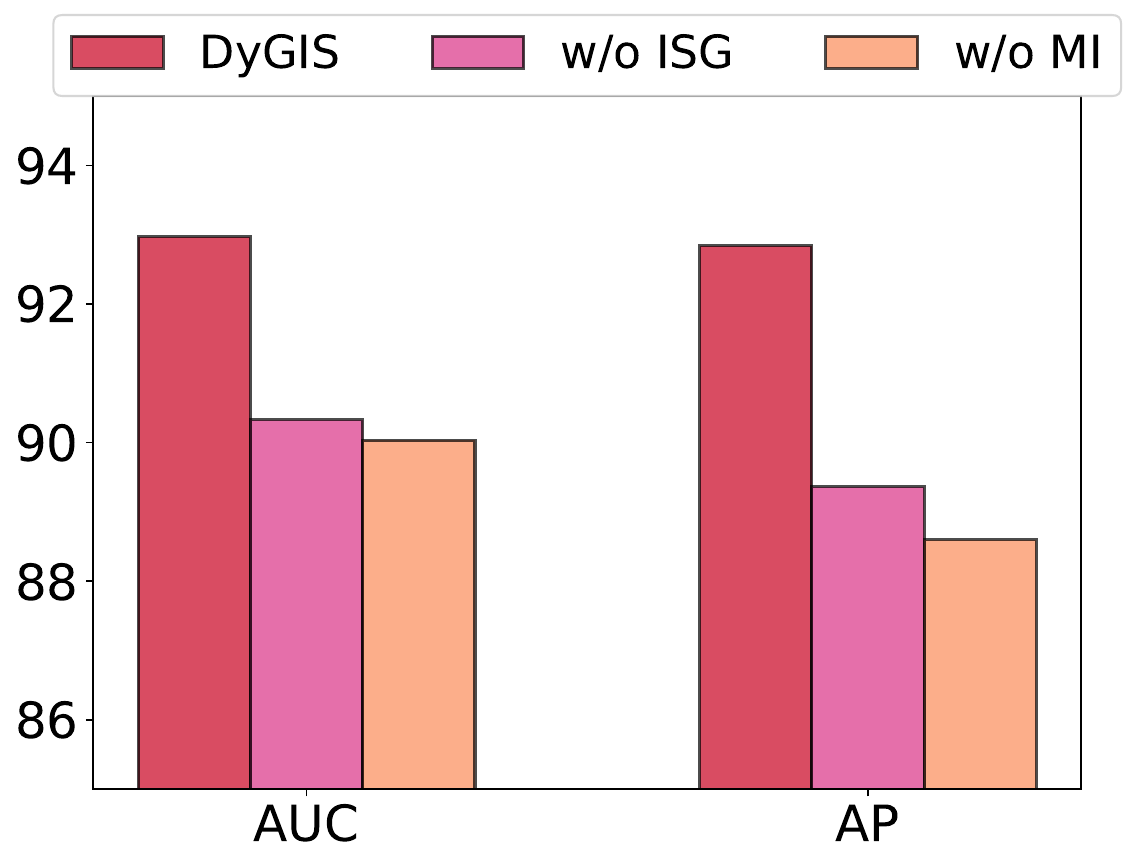}
}%
\subfloat[Email]{
\centering
\includegraphics[width=0.23\linewidth]{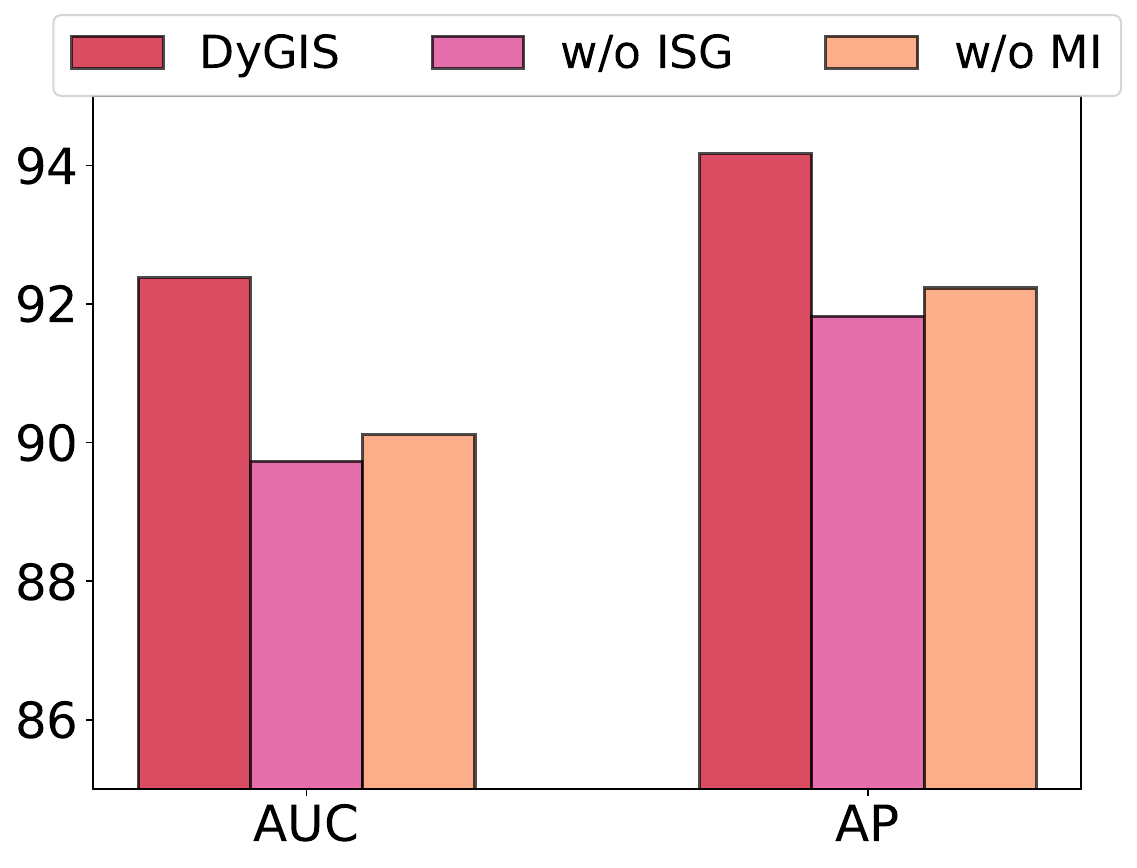}
}%
\subfloat[Socwiki]{
\centering
\includegraphics[width=0.23\linewidth]{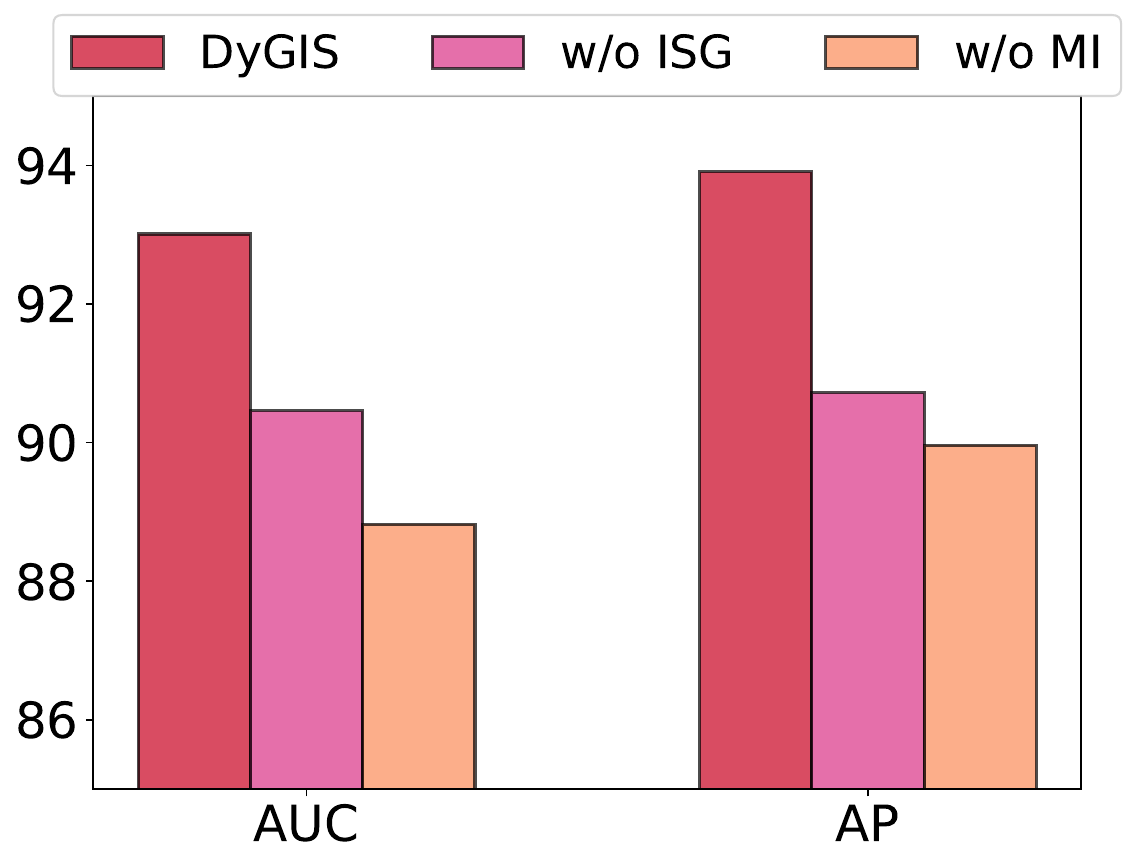}
}%
\centering
\caption{The ablation experiment results of four datasets on link detection task.}
\label{figure_ablation}
\end{figure*}

\begin{figure}[t]
\centering
\subfloat[Link Prediction]{
\centering
\includegraphics[width=0.48\linewidth]{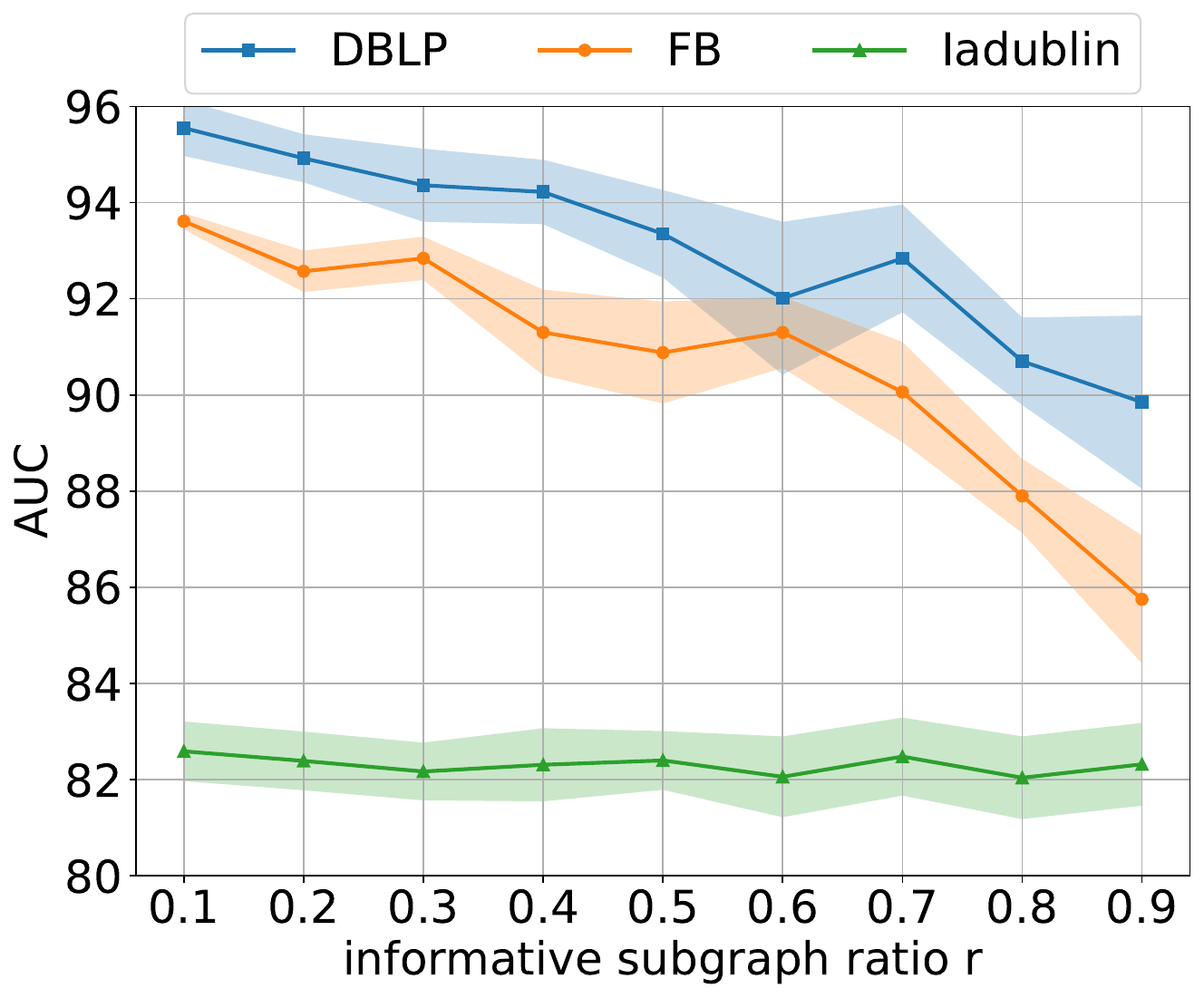}
}%
\subfloat[New Link Prediction]{
\centering
\includegraphics[width=0.48\linewidth]{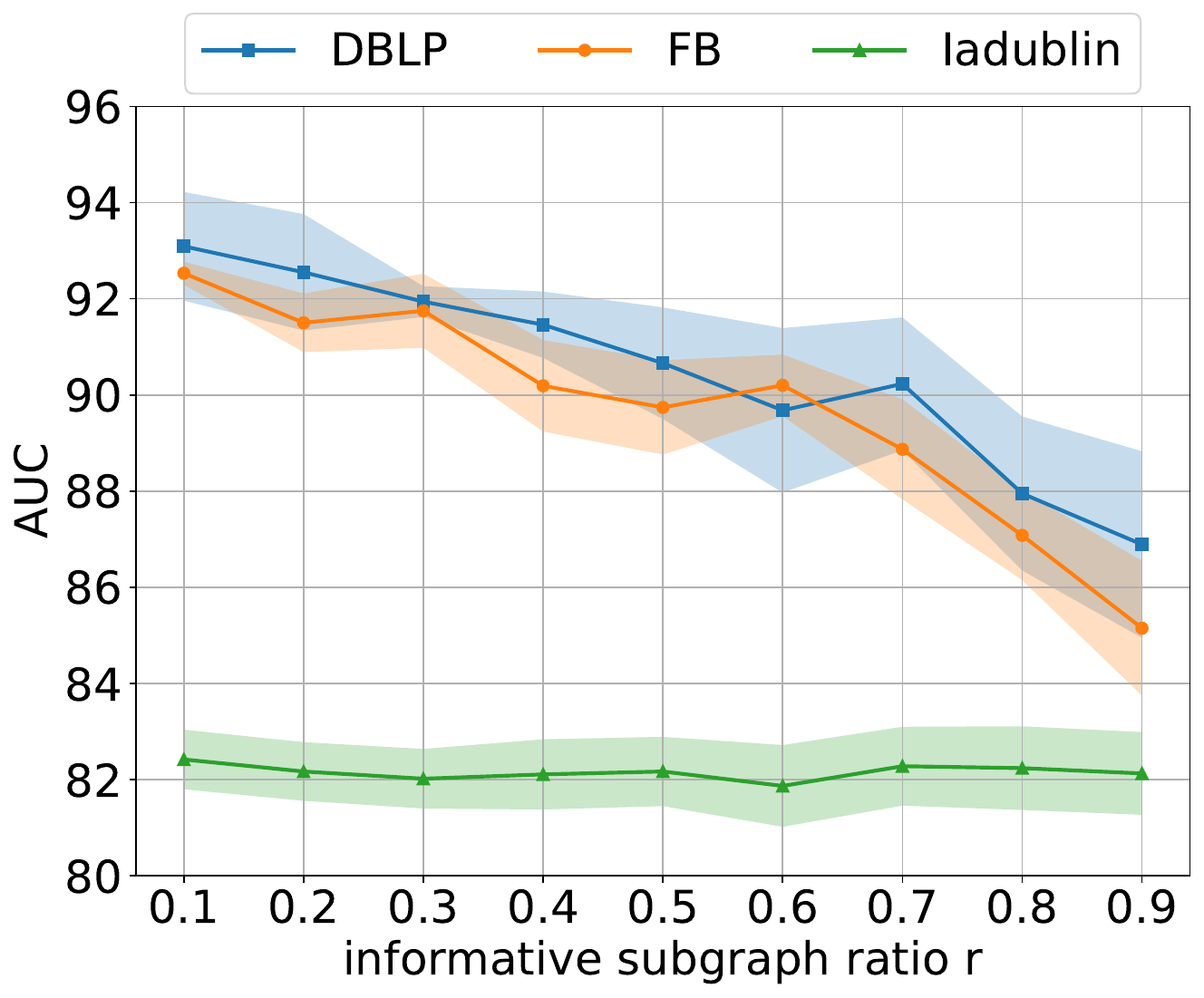}
}%
\centering
\caption{Parameter sensitivity analysis on link prediction and new link prediction task.}
\label{figure_param}
\end{figure}

\begin{figure}[t]
\centering
\includegraphics[width=0.5\columnwidth]{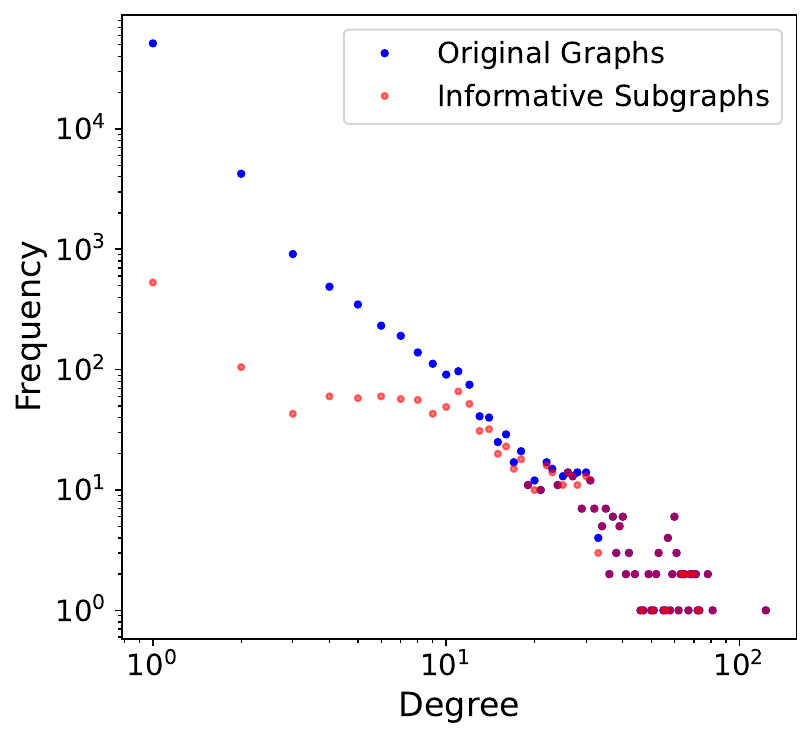} 
\caption{A case study on Email dataset with $r=0.1$. The blue points represent the degree distribution of the original dynamic graph, while the red points represent the degree distribution of the informative subgraph. Points with other colors indicate that the blue and red points overlap.}
\label{fig5}
\end{figure}

\subsection{Ablation Experiments} To further validate the effectiveness of each component of DyGIS, we conduct ablation experiments on variants of DyGIS. We name the DyGIS variants as follows:
\begin{itemize}
    \item w/o ISG: DyGIS without informative subgraph generator, \textit{i.e.}, we adopt a random masking strategy to obtain perturbed graphs as inputs and masked graphs as reconstruction targets for DGMAE.
    \item w/o MI: DyGIS without mutual information loss, \textit{i.e.}, DyGIS generates informative subgraphs without the constraint of mutual information loss.
\end{itemize}
We present the ablation experiment results in Figure \ref{figure_ablation}. We can observe that the performance decreases significantly after removing any of these two components. Specifically, the variant w/o ISG with a random masking strategy loses crucial information in dynamic graphs, resulting in a decrease in model performance. This also highlights that our strategy of generating informative subgraphs effectively preserves the crucial spatio-temporal information in dynamic graphs. Additionally, the performance of variant w/o MI drops explicitly as well. We analyze that the generated informative subgraphs miss some informative edges without the constraint of mutual information, leading to the loss of spatio-temporal information. This further validates that maximizing mutual information between the embedding of dynamic graphs and the embedding of noisy random graphs is an effective way to learn the informative structure of dynamic graphs. Overall, both of our components play crucial roles in enhancing the model's performance.


\subsection{Parameter Sensitivity Analysis} We further conduct parameter sensitivity analysis experiments to investigate the impact of key hyper-parameter informative subgraph ratio $r$ on model performance. Figure \ref{figure_param} shows the results for the link prediction and new link prediction tasks. For the DBLP and FB datasets, we observe the performance of the model decreases with increasing $r$. This can be connected with the information redundancy in dynamic graphs. The dynamic graph with a large number of redundant edges introduces a significant amount of redundant information to node representation. Utilizing a small number of informative subgraphs is more effective in obtaining meaningful node representation. In addition, the model's performance remains stable with the variation of $r$ in the Iadublin dataset. On the one hand, the Iadublin dataset may have relatively fewer redundant edges. On the other hand, the nodes and edges in the Iadublin dataset don't differ significantly in importance. Hence, the performance of DyGIS isn't sensitive to changes in $r$ in this dataset.

\subsection{Case Study} To explore whether our model truly generates informative subgraphs, we conduct a case study about degree distribution in the Email dataset and show results in Figure \ref{fig5}. It can be observed that the learned informative subgraphs include all nodes with large degrees, removing a significant number of trivial nodes and their connecting edges. In other words, DyGIS tends to identify nodes with large degrees and their edges to construct informative subgraphs. As the node degree decreases, DyGIS is more inclined to exclude that node and its connecting edges from the informative subgraph. In a graph, nodes with the highest degree typically play crucial hub roles and are named hub nodes. These nodes typically determine the evolution direction of the dynamic graph and have more meaningful information. Consequently, the informative subgraphs learned by our model indeed contain key nodes and edges in dynamic graphs, which leads to excellent performance even with a very small informative subgraph ratio ($r$=0.1).

\section{Conclusion}
In this paper, we propose a novel model, namely DyGIS, to apply MAE to dynamic graphs. To prevent the loss of informative spatio-temporal information in dynamic graphs, DyGIS generates informative subgraphs that guide the dynamic graph evolution to serve as the input for the dynamic masked autoencoder. Extensive experiments on multiple datasets validate the effectiveness of the proposed model across various tasks. Additionally, we conduct ablation experiments and parameter sensitivity analysis to verify the contributions of individual components in the model. Finally, we demonstrate the accuracy and effectiveness of the informative subgraphs identified through a case study on the Email dataset. In addition, our work also has some limitations. Currently, we mainly focus on generative SSL with masking on discrete dynamic graphs. In the future, we leave studying graph mask auto-encoders in continuous dynamic graphs and generalize our work to more challenging and meaningful tasks in more complex dynamic graph scenarios.

\section*{Acknowledgment}
This work was supported in part by the Zhejiang Provincial Natural Science Foundation of China under Grant LDT23F01012F01, in part by the National Natural Science Foundation of China under Grant 62372146, in part by the Fundamental Research Funds for the Provincial Universities of Zhejiang Grant GK229909299001-008, and in part by the HangzhouArtificial Intelligence Major Scientific and Technological In-novation Project under Grant 2022AIZD0114.
\bibliographystyle{IEEEtran}
\bibliography{ref}

\end{document}